\documentclass[letterpaper, 10 pt, conference]{ieeeconf}  

\IEEEoverridecommandlockouts                              

\usepackage[dvipsnames]{xcolor}
\usepackage{booktabs}
\usepackage{colortbl}
\usepackage{multirow}
\usepackage{graphicx}
\usepackage{xspace}
\usepackage{amssymb}
\usepackage{caption}
\usepackage{overpic}
\usepackage{url}
\usepackage[colorlinks=true, linkcolor=blue, citecolor=blue, urlcolor=blue]{hyperref}

\definecolor{firstcolor}{HTML}{BDE6CD}
\definecolor{secondcolor}{HTML}{E2EEBC}
\definecolor{thirdcolor}{HTML}{FFF8C5}

\newcommand{\fst}[1]{\cellcolor{firstcolor}\bfseries #1}
\newcommand{\snd}[1]{\cellcolor{secondcolor}#1}
\newcommand{\trd}[1]{\cellcolor{thirdcolor}#1}

\def\prjName{ProDyGS\xspace}

\title{\LARGE \bf
\prjName: Dynamic Gaussian Splatting \\ from a Single Static Monocular Camera
}

\author{Ugo Leone Cavalcanti$^{1}$ \hspace{1.4cm} Fabio Tosi$^{1}$ \hspace{1.4cm} Matteo Poggi$^{1}$ \hspace{1cm} Andrea Conti$^{2}$\\ Vladimir Zlokolica$^{2}$ \hspace{1cm} Valerio Cambareri$^{2}$ \hspace{1cm} Stefano Mattoccia$^{1}$ \vspace{0.2cm}\\%
$^{1}$Department of Computer Science and Engineering, University of Bologna, Italy \\
$^{2}$Sony Depthsensing Solutions, Brussels, Belgium \\
{\tt\small \textbf{Project page: \url{https://prodygs.github.io}} }
}

\begin{document}

\twocolumn[{
\renewcommand\twocolumn[1][]{#1}
\maketitle
\begin{center}
\vspace{-0.5cm}
\begin{overpic}
[width=\textwidth]{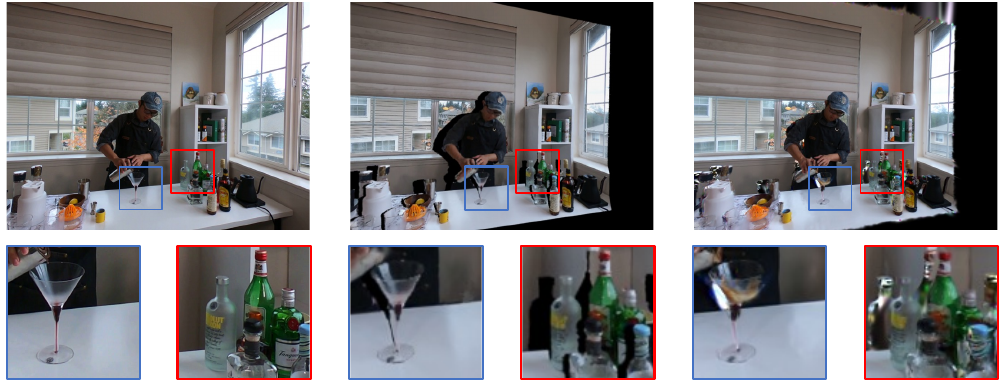}
\put(2, 36){\small \textcolor{yellow}{\texttt{\textbf{Ground Truth}}}}
\put(36, 36){\small \textcolor{yellow}{\texttt{\textbf{\prjName (ours)}}}}
\put(36, 34){\small \textcolor{yellow}{\texttt{\textbf{PSNR: 23.80}}}}
\put(70, 36){\small \textcolor{yellow}{\texttt{\textbf{MoDGS}}}}
\put(70, 34){\small \textcolor{yellow}{\texttt{\textbf{PSNR: 19.12}}}}
\end{overpic}
\end{center}\vspace{-0.3cm}
\captionof{figure}{\textbf{\prjName in action.} Our framework enables accurate 4D novel view synthesis from a single static camera.
\label{fig:teaser}
}
\vspace{1em}
}]

\begin{abstract}
We present \prjName, a novel dynamic 3D Gaussian Splatting framework for high-quality novel view synthesis from videos captured by a single static camera. While existing methods rely on multi-view setups or significant camera motion for geometric constraints, our approach addresses the challenging scenario where multi-view supervision is completely absent. We overcome this limitation by generating synthetic multi-view supervision through depth-guided proxy image synthesis. Specifically, we estimate temporally consistent depth maps using foundational monocular depth networks, then construct 3D Gaussian representations that generate proxy images from arbitrary viewpoints. A deformation network learns temporal dynamics by warping canonical Gaussians using this augmented supervision. Experiments on the DyNeRF dataset demonstrate that our method achieves state-of-the-art performance while requiring only monocular depth estimation as external supervision, outperforming approaches that rely on stronger priors such as scene flow.
\end{abstract}    
\section{Introduction}

The Novel View Synthesis task (NVS) lies at the intersection of computer vision and graphics, playing a crucial role in immersive applications such as augmented and virtual reality. 
It has gained increasing popularity in the twenties, thanks to the advent of recent approaches such as Neural Radiance Fields (NeRF) \cite{mildenhall2020nerf} and 3D Gaussian Splatting (3DGS) \cite{kerbl20233d}, which accurately model the 3D geometry of a single, static scene and enable rendering from new, arbitrary viewpoints.
Both frameworks and their derivatives have deep roots in multi-view  geometry, ensuring unprecedented photorealism when the input images used for training are available in large quantities and have been collected densely throughout the scene. 
Conversely, the quality of the rendering process rapidly deteriorates when the input images are insufficient, due to the weaker multi-view  geometry guidance available during training -- which needs to be compensated through the use of auxiliary priors \cite{kangle2021dsnerf,chung2024depth}.

The success of NeRF and 3DGS also led the community to extend NVS to the more challenging Dynamic View Synthesis task (DVS), where the temporal axis lifts 3D geometry into a 4D field, enabling rendering of the scene both from arbitrary viewpoints and at any desired timeframe. This process is usually simplified by learning a canonical 3D representation of the scene, either in the form of a NeRF or 3DGS, and a \textit{deformation field} \cite{park2021nerfies} predicted by a shallow network and applied to the canonical representation to warp the actual scene elements to any desired timeframe.
Accordingly, the availability of multi-view, synchronized videos densely distributed throughout the scene \cite{li2022neural} is crucial for effectively transitioning from NVS to DVS while retaining the highest photorealism. Nonetheless, some methods yield high-quality DVS even when processing a monocular video, if the camera performs sufficiently large movements -- thus providing adequate multi-view  geometry guidance for training.
However, such large camera motion -- also referred to as \textit{teleporting camera motion} -- is rarely observed in casual videos \cite{gao2022monocular}, which are typically captured through slow and smooth trajectories rather than sudden viewpoint changes.

The most challenging scenario occurs when the video is captured by a static monocular camera: in such a case, the complete absence of multi-view geometry guidance causes existing DVS frameworks to fail at rendering high-quality novel views. Although some works overcome this lack of supervision with external priors, focusing on modeling either the structure or the temporal dynamics in the video -- e.g., by relying on pre-computed depth, optical flow or scene flow priors \cite{qingming2025modgs} -- we argue that they overlook the real problem, which is the lack of multi-view geometry constraints to guide the training process from different viewpoints. 

In this paper, we attack the problem directly from this perspective and aim to partially restore this paramount supervision that is completely absent in videos captured by a single static camera. 
We present \textbf{\texttt{\prjName}}, a dynamic Gaussian Splatting framework optimized by \textit{inducing} multi-view geometry supervision. Specifically, given a monocular video collected by a static camera, we estimate a temporally consistent depth map for each frame. This is used to generate a set of \textit{proxy images} through forward warping, according to arbitrary viewpoints sampled around the original camera position. These proxy images are then added to the training set and used to supervise our model, implemented with a 3DGS backbone and a deformable network. Despite its simplicity, our experiments on the DyNeRF dataset \cite{li2022neural} highlight the superior effectiveness of our solution over existing methods that exploit even stronger priors, such as scene flow \cite{qingming2025modgs} -- see Fig.~\ref{fig:teaser} for a qualitative example.

Our main contributions can be summarized as follows:

\begin{itemize}
    \item We introduce \prjName, a novel dynamic Gaussian Splatting framework tailored for monocular videos captured by a static camera.

    \item \prjName overcomes the lack of multi-view geometry guidance by generating proxy images from arbitrary viewpoints. This is achieved through forward warping, after estimating the depth of each frame in the video.

    \item \prjName achieves state-of-the-art performance with minimal requirements in terms of external priors. 
\end{itemize}

\section{Related Work}

We review existing literature relevant to our work, spanning dynamic neural radiance fields, dynamic 3D Gaussian splatting methods, and depth priors for constrained scenarios.

\textbf{Dynamic NeRF.} Neural Radiance Fields (NeRF) were introduced for novel view synthesis of static scenes \cite{mildenhall2020nerf} and have since become a cornerstone in implicit 3D scene representation, as reviewed in a comprehensive survey \cite{gao2022nerf,irshad2024neural,rabby2023beyondpixels}. These methods have been extended to dynamic scenarios by incorporating temporal inputs and deformation fields, as in D-NeRF \cite{pumarola2021d} and Nerfies \cite{park2021nerfies}, with HyperNeRF \cite{park2021hypernerf} further handling topological changes. Other methods combine static and dynamic radiance fields \cite{gao2021dynamic} or decouple moving objects from the background in a self-supervised manner \cite{wu2022d}. To improve efficiency, explicit and hybrid representations such as \cite{wang2022fourier, fang2022fast, cao2023hexplane, fridovich2023k, wang2023mixed, shao2023tensor4d, song2023nerfplayer} reduce training time and memory usage while maintaining quality. Recent works also include motion-aware volumetric image-based rendering for long dynamic videos \cite{li2023dynibar}, forward flow warping \cite{guo2023forward}, and ray-conditioned sampling for real-time 6-DoF video \cite{attal2023hyperreel}, with robustness to challenging camera poses addressed in \cite{liu2023robust}.


\textbf{Dynamic 3D Gaussian Splatting.} Following the evolution of NeRF from static to dynamic scenes, 3DGS has progressed from its original static representation \cite{kerbl20233d} to handle temporal dynamics. For a broader overview of 3DGS, see recent surveys \cite{fei20243d,chen2024survey,wu2024recent,tosi2024nerfs,rabby2023beyondpixels}. Early works \cite{yang2023gs4d, wu20244d, lin2024gaussian, luiten2024dynamic} introduced key methods using time-dependent parameters, deformation fields, and persistent tracking. Subsequent methods have focused on efficiency and compactness \cite{katsumata2023efficient, katsumata2024compact, kratimenos2024dynmf, huang2024sc}, motion-aware representations \cite{guo2024motion, park2025splinegs}, and controllable editing \cite{kwon2025efficient}. Advanced techniques include specialized temporal strategies \cite{shaw2024swings, li2024spacetime, yan20244d, lee2024fully}, deformable representations \cite{yang2024deformable, bae2024per, lu20243d, das2024neural}, streaming and real-time approaches \cite{sun20243dgstream, yan2025instant, wang2025freetimegs}, and novel formulations such as 4D rotor Gaussians \cite{duan20244d}, semantic integration \cite{labe2024dgd}, motion scaffolds \cite{lei2025mosca} and so on \cite{stearns2024dynamic, liang2025gaufre, lu2025bard, zhu2025voxelsplat, kwak2025modec}. However, most existing methods assume either multi-view setups or monocular videos with substantial camera motion to establish sufficient geometric constraints for reconstruction. Only MoDGS \cite{qingming2025modgs} specifically addresses the challenging scenarios of casually-captured monocular videos with static or slowly-moving cameras, using depth priors to compensate for weak geometric constraints. Similarly, our approach tackles the particularly restrained setting of static monocular cameras for dynamic reconstructions, representing an even more challenging scenario that remains largely unexplored by other methods.   

\begin{figure*}[t]
    \centering
    \includegraphics[width=0.95\textwidth]{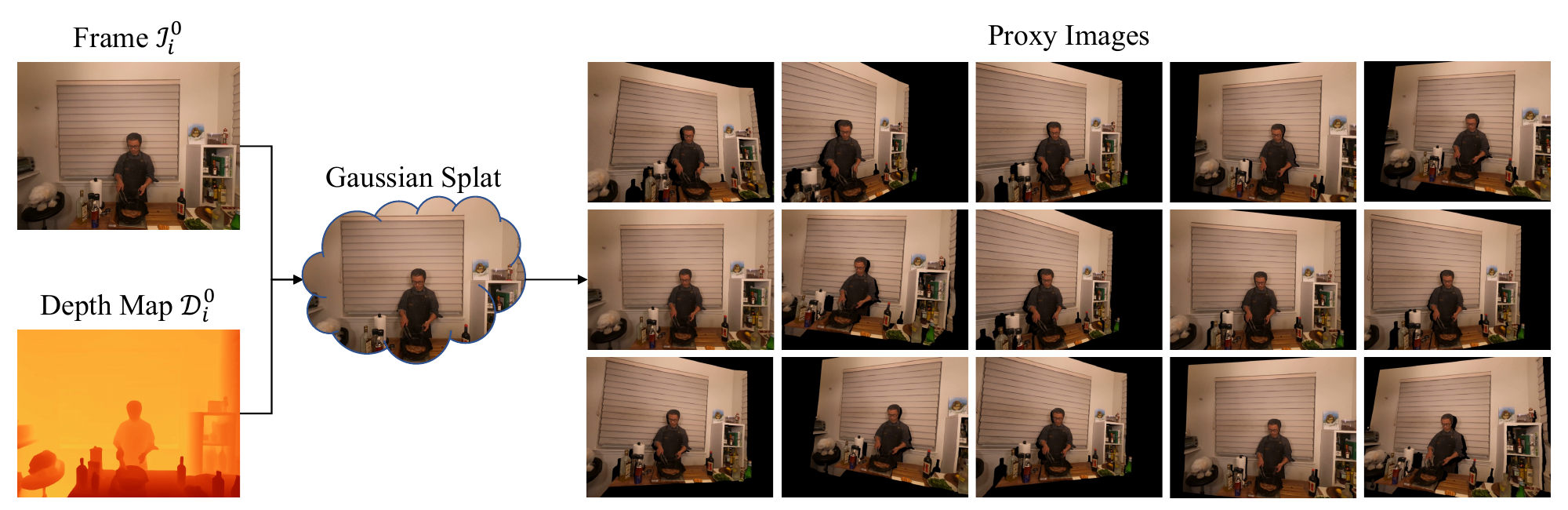}
    \caption{\textbf{Proxy images generation.} In \prjName for each frame $\mathbf{I}_i^0$ of the input sequence we initialize a lightweight 3DGS model out of each pair $(\mathbf{I}_i^0, \dot\mathbf{D}_i^0)$. We then use the 3DGS model to render the additional proxy images $\{ \{ \mathbf{I}_i^j \}_{j=1}^M \}_{i=0}^N$ from each of $M$ viewpoints $\{ \mathbf{\Pi}^j \}_{j=1}^M$, by means of differentiable splatting $\mathcal{S}(\mathbf{\Pi}^j,\mathcal{G}_i)$.}
    \label{fig:proxyGenImg}
    \vspace{-1em}
\end{figure*}

\textbf{Depth Estimation for Scene Priors.} Recent advances in monocular depth estimation have been driven by foundation models trained on massive datasets, achieving remarkable generalization across diverse scenarios. Early deep learning approaches~\cite{Eigen2015,Ranftl2020,Bhat2021,Fu2018} evolved through architectural improvements and advanced loss formulations, while self-supervised methods~\cite{Zhou2017,Godard2017} leveraged photometric consistency for scalability. The field has shifted toward large-scale multi-dataset training strategies with foundation models like MiDaS~\cite{Ranftl2020,Ranftl2021}, transformer-based architectures~\cite{Ranftl2021,Bhat2023}, and massive-scale approaches like Depth Anything~\cite{Yang2024,Yang2024v2} trained on over 60 million images. Recent developments include diffusion-based methods~\cite{ke2024repurposing,Fu2024,Ji2023}, metric depth estimation frameworks~\cite{Bhat2023,yin2023metric3d,hu2024metric3d,piccinelli2024unidepth,piccinelli2025unidepthv2,bochkovskii2024depth}, temporal consistency approaches~\cite{shao2025chronodepth,hu2024depthcrafter,ke2024rollingdepth,chen2025video,yang2024depth,lu2025align3r}, and refinement techniques~\cite{aleotti2021neural,tosi2024neural,zhang2024betterdepth} that enhance edge sharpness and reduce artifacts. These robust, generalizable depth estimation networks provide crucial geometric priors for dynamic scene reconstruction, particularly in constrained scenarios where traditional multiview constraints are unavailable. In our framework, we leverage these advances to obtain reliable depth information that compensates for the lack of geometric constraints in static monocular camera setups.
\section{Methodology}

Our proposed framework, \prjName, addresses the challenge of reconstructing dynamic 3D scenes from videos captured by a single, quasi-static camera. This is achieved by leveraging depth as the central modality, which is recovered for each frame by deploying an off-the-shelf single-image depth estimation model and used to induce additional multi-view supervision by generating a set of proxy images from arbitrary viewpoints around the real camera.

\subsection{Depth Estimation and Refinement} \label{depth_refinement}

As the first step in our pipeline, we aim to recover the 3D structure of the scene observed through a single static camera. For this purpose, we use a pre-trained single-image depth estimator $\mathbf{\Phi}$ to predict a depth map $\mathbf{D}_i$ for each image $\mathbf{I}_i^0$ in the video sequence $\{ \mathbf{I}_i^0 \}_{i=0}^N $:

\begin{equation}
    \{ \mathbf{D}_i^0 \}_{i=0}^N = \{ \mathbf{\Phi} (\mathbf{I}_i^0) \}_{i=0}^N  
\end{equation}
Specifically, we select Depth Pro \cite{bochkovskii2024depth} for its fast inference speed and well edge-aligned, though over-smoothed, depth maps even for high-resolution images. Moreover, we post-process the estimated depth maps $\{ \mathbf{D}_i^0 \}_{i=0}^N$ to improve both single-frame accuracy and temporal consistency.

\textbf{Neural Refinement.} The availability of both accurate and high-resolution depth maps is crucial for the subsequent stages performed by \prjName. Therefore, we leverage a second, off-the-shelf model $\mathbf{\Psi}$ to refine the preliminary depth maps estimated by the depth estimator. We deploy the Neural Disparity Refinement framework \cite{aleotti2021neural,tosi2024neural}, which is specifically designed to enhance both depth accuracy and sharpness, thanks to its continuous output formulation.

Given the set of paired images and initial depth maps $\{ (\mathbf{I}_i^0, \mathbf{D}_i^0) \}_{i=0}^N$, we forward each pair to $\mathbf{\Psi}$ to obtain a set of new depth maps $\{\hat\mathbf{D}_i^0\}_{i=0}^N$ 

\begin{equation}
    \{ \hat\mathbf{D}_i^0 \}_{i=0}^N = \{ \mathbf{\Psi} (\mathbf{I}_i^0, \mathbf{D}_i^0) \}_{i=0}^N  
\end{equation}

Improving the quality of estimated depth maps provides both a more reliable basis for constructing 3D Gaussians, as well as enhancing the proxy image generation, resulting in more robust training signals for the overall framework.

\textbf{Temporal Consistency.} 
A well-known limitation of monocular depth estimation networks is their tendency to exhibit limited temporal consistency across a video sequence in a frame-to-frame fashion \cite{shao2025chronodepth}. Such inconsistency often manifests as scale drift and can significantly degrade the quality of the final reconstruction in our framework, as temporal depth fluctuations cause scene objects to warp or distort over time. Therefore, compensating for this drift within \prjName is crucial to achieve optimal results. 

When capturing videos with a static camera, we argue that a significant portion of the scene maintains the same depth over time -- i.e., background regions -- and can be used to enforce scale consistency across the entire sequence. 
Therefore, we select the initial frame $\mathbf{I}_0$ and aim to align the remaining frames with it: for this purpose, we identify a background mask $\mathbf{B}_i^0$ in each refined depth map $\hat\mathbf{D}_i^0$ by examining the predicted optical flow between each frame $\mathbf{I}_i^0$ and $\mathbf{I}_0$, and compute scale and shift factors $(\alpha_i^0,\beta_i^0)$ between $\hat\mathbf{D}_i^0$ and the reference $\hat\mathbf{D}_0$ on static regions only:

\begin{equation}
    (\alpha_i,\beta_i) = \mathop{argmin}_{\alpha,\beta} \left( \alpha \mathbf{B}_i^0\hat\mathbf{D}_i^0 + \beta - \mathbf{B}_i^0\hat\mathbf{D}_0 \right)^2
\end{equation}
and then obtained scale-aligned depth maps $\{ \dot\mathbf{D}_i^0 \}_{i=1}^N$ as 

\begin{equation}
    \{ \dot\mathbf{D}_i^0 \}_{i=1}^N = \{ \alpha_i \hat\mathbf{D}_i^0 + \beta_i \}_{i=1}^N  
\end{equation}

This simple yet effective alignment step mitigates scale drift and plays a crucial role in maintaining structural stability during \prjName optimization.
In our experiments, we computed the optical flow analysis to separate static background regions from dynamic scene elements using the off-the-shelf SEA-RAFT \cite{wang2024searaftsimpleefficientaccurate} model.

\begin{figure*}[t]
    \centering
    \includegraphics[width=0.95\textwidth]{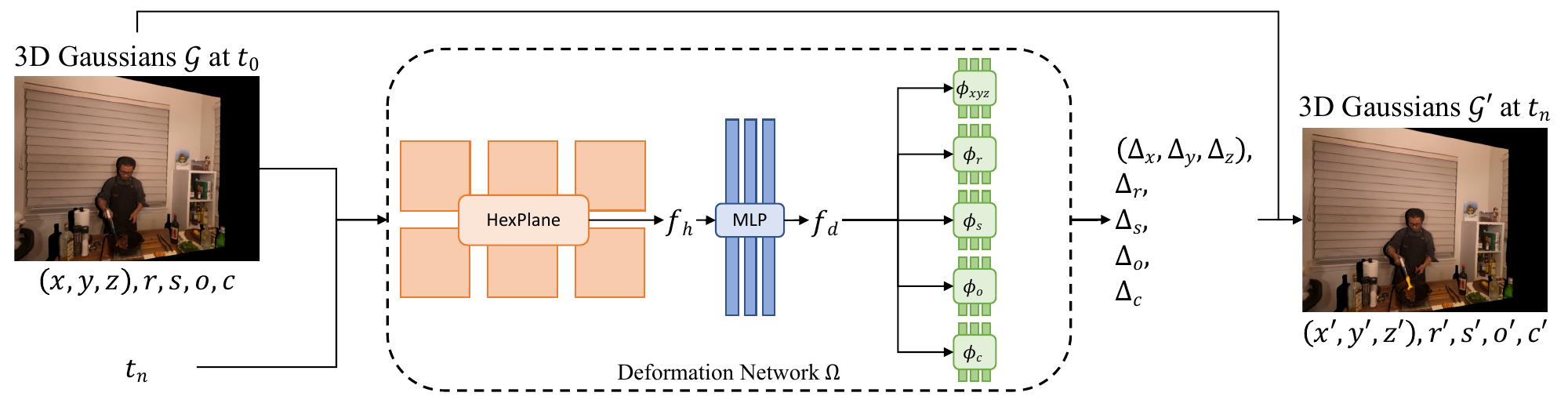}
    
    \caption{\textbf{Overview of our \prjName framework.} The deformation network $\Omega$ consists of a spatio-temporal encoder, based on a multi-resolution HexPlane and a lightweight MLP, that extracts features from both the Gaussians and the timestamp. A multi-head decoder $D = \{\phi_x, \phi_s, \phi_r, \phi_o, \phi_c\}$ then predicts per-Gaussian updates of position, scale, rotation, opacity, and color. These updates are applied to the canonical Gaussians $\mathcal{G}$ to obtain the deformed Gaussians $\mathcal{G}^\prime_i$}
    \label{fig:architectureImg}

\end{figure*}
\subsection{Proxy Images Generation}

Once accurate depth maps are available for each frame of the sequence, we can use them to retrieve the missing multi-view geometry cues necessary for optimizing \prjName, as
illustrated in Fig.~\ref{fig:proxyGenImg}. 
Purposely, we fix the real camera pose $\mathbf{\Pi}^0$ -- e.g., to identity, or to an arbitrary pose -- and sample a set of camera poses $\{ \mathbf{\Pi}^j \}_{j=1}^M$ arranged in a frontal hemispherical configuration around the scene. Then, for each frame in the original sequence $\{ \mathbf{I}_i^0 \}_{i=0}^N$, we aim to generate a new set of proxy images, resulting in a new set $\{ \{ \mathbf{I}_i^j \}_{j=0}^M \}_{i=0}^N$ of multi-view sequences. 

This is achieved by initializing a lightweight 3DGS model from each pair $(\mathbf{I}_i^0, \dot\mathbf{D}_i^0)$, where Gaussians $\mathcal{G}_i = \{g_0, g_1, ... g_G \}_i$ are defined as:
\begin{equation}
    g_k(\mathbf{x}) = e^{-\frac{1}{2} (\mathbf{x}-\boldmath{\mu}_k)^\top \mathbf{\Sigma}_k^{-1} (\mathbf{x}-\boldmath{\mu}_k)}
\end{equation}
by means of a center position $\boldmath{\mu}_i \in \mathbb{R}^3$, a covariance matrix $\mathbf{\Sigma}_i \in \mathbb{R}^{3 \times 3}$, in addition to opacity $o_i \in [0,1]$, and a view-dependent colors $\mathbf{c}_i$ represented using spherical harmonics.
Gaussians are seeded by setting $\mu_k$ as the 3D coordinate of each pixel in $\dot\mathbf{D}_i^0$ and their color is initialized according to $\mathbf{I}_i^0$. These 3DGS models are then optimized for as few as 100 steps and used to render the additional proxy images $\{ \{ \mathbf{I}_i^j \}_{j=1}^M \}_{i=0}^N$ from viewpoints $\{ \mathbf{\Pi}^j \}_{j=1}^M$, through differentiable splatting $\mathcal{S}(\mathbf{\Pi}^j,\mathcal{G}_i)$ \cite{kerbl20233d}. 
This is achieved by projecting 3D Gaussians onto a 2D image plane --  projecting the 3D covariance into 2D as $\mathbf{\Sigma}' = \mathbf{JW \Sigma W^T J^T}$ and the center as $\boldmath{\mu}' = \mathbf{JW}\boldmath{\mu}$, and by merging the colors  $C$ of overlapping 3D Gaussian splats into pixels, sorted by depth:

\begin{equation}
C = \sum_{i \in \mathcal{N}} \mathbf{c}_i \alpha_i \prod_{j=1}^{i-1} (1 - \alpha_j) \quad\quad\quad 
\end{equation}
with $\alpha_i := o_i \exp\left(-\frac{1}{2} (\mathbf{x}' - \boldmath{\mu}'_i)^\top {\Sigma}'^{-1}_i (\mathbf{x}' - \boldmath{\mu}'_i)\right)$.

The lightweight 3DGS models are then discarded, with only the sampled poses and generated proxy images being retained and used to optimize \prjName. 

\begin{table*}[t]
    \centering
    \caption{\textbf{Per-scene results on the DyNeRF dataset~\cite{li2022neural}.} We compare our method with  HexPlane~\cite{cao2023hexplane}, SC-GS~\cite{huang2024sc}, Deformable GS (D-GS)~\cite{yang2024deformable} and MoDGS~\cite{qingming2025modgs} in PSNR$\uparrow$, SSIM$\uparrow$, and LPIPS$\downarrow$.}
    \label{tab:DyNeRF_perscene}
    \renewcommand{\tabcolsep}{12pt}
    \resizebox{1.0\linewidth}{!}{
    \begin{tabular}{l|ccc|ccc|ccc}
        \toprule
        \textbf{Method} & \multicolumn{3}{c|}{\textbf{\texttt{cut\_roasted\_beef}}} & \multicolumn{3}{c|}{\textbf{\texttt{sear\_steak}}} & \multicolumn{3}{c}{\textbf{\texttt{coffee\_martini}}} \\
        \cmidrule(lr){2-4} \cmidrule(lr){5-7} \cmidrule(lr){8-10}
                        & PSNR & SSIM & LPIPS & PSNR & SSIM & LPIPS & PSNR & SSIM & LPIPS \\
        \midrule
        {HexPlane~\cite{cao2023hexplane}}  & 16.76 & 0.5382 & 0.5054 & 16.89 & 0.5897 & 0.5049 & 13.26  & 0.4049 & 0.5835 \\
        {SC-GS~\cite{huang2024sc}}     & 20.69 & 0.7414 & 0.2625 & 21.23 & 0.7870 & 0.2188 & 19.02  & 0.7124 & 0.2151 \\
        {D-GS~\cite{yang2024deformable}}      & \trd 22.20 & \trd 0.7808 & \trd 0.1931 & \snd 23.56 & \trd 0.8101 & \trd 0.1773 & \trd 19.23  & \trd 0.7013 & \trd 0.2270 \\
        {MoDGS~\cite{qingming2025modgs}}      & \snd 23.98 & \snd 0.8221 & \snd 0.1438 & \trd 23.53 & \snd 0.8126 & \snd 0.1642 & \snd 21.37  & \snd 0.7962 & \snd 0.1473 \\
        \textbf{\prjName (ours)} & \fst 27.55 & \fst 0.8986 &  \fst 0.1010 & \fst 29.57 & \fst 0.9220 & \fst 0.1069 & \fst 22.73  & \fst 0.8201 & \fst 0.1344 \\
        \bottomrule
    \end{tabular}
    }\vspace{0.1cm}

\resizebox{1.0\linewidth}{!}{
    \begin{tabular}{l|ccc|ccc|ccc}
        \toprule
        \textbf{Method} & \multicolumn{3}{c|}{\textbf{\texttt{cook\_spinach}}} & \multicolumn{3}{c|}{\textbf{\texttt{flame\_steak}}} & \multicolumn{3}{c}{\textbf{\texttt{flame\_salmon\_1}}} \\
        \cmidrule(lr){2-4} \cmidrule(lr){5-7} \cmidrule(lr){8-10}
                        & PSNR & SSIM & LPIPS & PSNR & SSIM & LPIPS & PSNR & SSIM & LPIPS \\
        \midrule
        {HexPlane~\cite{cao2023hexplane}}  & 16.95 & 0.7286 & 0.2223 & 16.97 & 0.7528 & 0.2543 & 11.16  & 0.3417 & 0.6382 \\
        {SC-GS~\cite{huang2024sc}}     & 16.70 & 0.7377 & 0.2117 & 17.31 & 0.7532 & 0.2527 & 17.65  & 0.6834 & 0.2253 \\
        {D-GS~\cite{yang2024deformable}}      & \trd 17.20 & \trd 0.7195 & \trd 0.2329 & \trd 16.62 & \trd 0.7523 & \trd 0.2559 & \trd 18.48  & \trd 0.7038 & \trd 0.2166 \\
        MoDGS~\cite{qingming2025modgs}      & \snd 22.40 & \snd 0.7823 & \snd 0.1728 & \snd 23.23 & \snd 0.8083 & \snd 0.1592 & \snd 21.33  & \snd 0.8038 & \snd 0.1399 \\
        \textbf{\prjName (ours)} & \fst 27.80 & \fst 0.9000 & \fst 0.1038 & \fst 26.31 & \fst 0.9176 & \fst 0.0933   & \fst 23.20 & \fst 0.8228 & \fst 0.1400\\
        \bottomrule
    \end{tabular}

    }
\vspace{-1em}
\end{table*}

\subsection{\prjName Framework}

Following 4D Gaussian Splatting \cite{wu20244d}, we implement our \prjName by combining a set of 3D Gaussians $\mathcal{G}$, assumed to represent the scene in an arbitrary canonical space, together with a deformation network $\mathbf{\Omega}$ (see Figure~\ref{fig:architectureImg} for an overview of the full pipeline). Given a camera pose matrix $\mathbf{\Pi}^j$ and a timestamp $i$, a novel view $\hat{\mathbf{I}^j_i}$ is rendered through differentiable splatting, formulated as $\hat{\mathbf{I}^j_i} = \mathcal{S}(\mathbf{\Pi}^j,\mathcal{G}^\prime_i)$, with $\mathcal{G}^\prime_i$ being the deformed Gaussians defined as: 

\begin{equation}\label{eq:gaussians}
    \mathcal{G}^\prime_i = \mathcal{G} + \Delta_i \mathcal{G}
\end{equation}
with $\Delta_i \mathcal{G}$ being the deformation field, predicted by $\mathbf{\Omega}$ as $\Delta_i \mathcal{G} = \mathbf{\Omega}( \mathcal{G}, i)$
modeling the temporal dynamics of the scene.

Specifically, $\mathbf{\Omega}$ consists of a spatio-temporal encoder $H$ that extracts features from both the Gaussians and the timestamp, followed by a multi-head Gaussian deformation decoder $D$ that predicts per-Gaussian updates. 

These updates allow \prjName to project the original Gaussians $\mathcal{G}$ from the canonical space into the real 4D space,
enabling dynamic novel-view synthesis with minimal overhead while preserving the differentiability of the splatting operation.

\textbf{Spatio-Temporal Encoder.} Inspired by \cite{wu20244d}, we assume that nearby 3D Gaussians share
similar spatio-temporal properties, such that they can be grouped within spatial plane bounding voxels while their deformations can also be  encoded in nearby temporal voxels. We therefore design an encoder $H$ that combines a multi-resolution HexPlane representation \cite{cao2023hexplane} with a lightweight MLP $\phi_d$ to efficiently extract Gaussian features.


Specifically, we use a multi-resolution HexPlane representation \cite{fridovich2023k}, composed of six plane modules $R_l(i, j) \in \mathbb{R}^{h\times l N_i \times l N_j}$, where $h$ denotes the hidden dimension of features, $N$ is the basic resolution of the voxel grid, and each module corresponds to one of the spatial or spatio-temporal planes $(x,y)$, $(x,z)$, $(y,z)$, $(x,t)$, $(y,t)$, and $(z,t)$, operating at two spatial scales ($l \in \{1,2\}$).

For each 4D coordinates set, 
we extract separate voxel features $f_h \in \mathbb{R}^{hl}$, that are then concatenated 
and forwarded shallow MLP $\phi_d$ to predict the final features in $f_d = \phi_d(f_h)$.

\textbf{Multi-Head Decoder.}
We then introduce a multi-head Gaussian deformation decoder 
$D$, defined by five shallow MLPs $\{\phi_x, \phi_s, \phi_r, \phi_o, \phi_c\}$ responsible for estimating per-Gaussian updates at each timestamp $i$ from the encoded features $f_h$. 
Specifically, the five branches are responsible for estimating the temporal deformation of position $(\Delta_x = \phi_x)$, scale $(\Delta_s = \phi_s)$, rotation $(\Delta_r = \phi_r)$, opacity $(\Delta_o = \phi_o)$, and color $(\Delta_c = \phi_c)$.
The five outputs are finally added to the canonical Gaussians $\mathcal{G}$ to obtain $\mathcal{G}^\prime_i$ as defined in Eq. \ref{eq:gaussians}, which is then used for differentiable splatting.


\subsection{Loss function}

The training of our deformation network is guided by a composite loss with three components:

\begin{equation}
    \mathcal{L} = \mathcal{L}_{rgb}
    + \mathcal{L}_{depth}
    + \mathcal{L}_{reg} .
\end{equation}

The RGB loss combines L1 distance and SSIM, and is applied only to visible regions of the rendered image, focusing supervision where the splat provides valid predictions:
\begin{equation}
    \mathcal{L}_{rgb} = |\hat{\mathbf{I}} - \mathbf{I}| + \gamma \left(1 - SSIM(\hat{\mathbf{I}}, \mathbf{I})\right).
\end{equation}

The depth loss follows the scale- and shift-invariant (SSI) formulation \cite{Ranftl2020} comparing the predicted depth ($\hat{\mathbf{D}}$) with the refined and temporally aligned proxy depth ($\dot{\mathbf{D}}^0$):
\begin{equation}
    \mathcal{L}_{depth} = \frac{1}{2M} \sum_{i=1}^{M} \rho\Big(\hat{\mathbf{D}}_i - \dot{\mathbf{D}}_i^0\Big),
\end{equation}

where $M$is the number of valid pixels and $\rho$ is a robust loss 
(i.e., L2 loss \cite{Ranftl2020}),
comparing scaled and shifted predictions with proxy depth maps to enforce geometrically consistent deformations despite global depth ambiguities.

Finally, the regularization term $\mathcal{L}_{reg} = \mathcal{L}_{TV}$ applies grid-based total variation to the HexPlane feature representations, following~\cite{wu20244d}, to discourage degenerate solutions and promote stable training; 

 $\mathcal{L}_{TV}$ denotes grid-based total variation regularization to the multi-resolution HexPlane feature grids following \cite{cao2023hexplane,fridovich2023k,sun2022direct}, which promotes smooth gradients and stable training of the deformation network.

\section{Experiments}

\subsection{Implementation Details}
Our framework is implemented in PyTorch. The deformation network is trained for 40k steps using the Adam optimizer, initialized with a learning rate of $1\times 10^{-4}$. Training converges within approximately three hours on a single NVIDIA RTX 5090 GPU. 


\subsection{Evaluation Protocol}

\textbf{Datasets.}
Our primary benchmark is the DyNeRF dataset \cite{DBLP:journals/corr/abs-2103-02597}, which contains six dynamic scenes recorded with 18-20 synchronized cameras over 10-30 seconds, featuring tabletop activities with accurate camera pose annotations. 
The choice of this dataset is dictated by the requirements of our evaluation protocol. 
In particular, the proposed evaluation assumes (i) multiple synchronized videos acquired from different viewpoints, (ii) static cameras with known calibration parameters, and (iii) sufficiently dense temporal sampling. 

Only a limited number of publicly available datasets simultaneously satisfy these constraints. Among them, DyNeRF has emerged as one of the most widely adopted benchmarks since it provides densely sampled multiview videos without temporal downsampling, making it particularly suitable for evaluating motion smoothness and temporal coherence in dynamic NVS. 
In all experiments, camera 0 is used for training, while cameras 5 and 6 serve as novel view evaluation targets.

To prove the robustness and generalization of the proposed approach beyond the above benchmark, we perform some experiments on sequences from the Light Field Video dataset \cite{broxton2020immersive}. 
Since this dataset differs from DyNeRF in camera configuration, scene layout, and capture setup, it provides a complementary test scenario for qualitatively evaluating the behavior of the proposed system under different acquisition conditions.

\textbf{Evaluation Setting.} Following MoDGS \cite{qingming2025modgs}, 
we use a single real view (camera 0) and synthetically render the remaining viewpoints using their ground-truth poses. The deformation network is trained on these synthetic frames from all cameras except 5 and 6, which are held out for testing. At test time, we render novel views at the poses of cameras 5 and 6 and compare them against the corresponding ground-truth captures, providing a fair measure of generalization to unseen viewpoints.

\begin{table}[t]
\centering
\renewcommand{\tabcolsep}{12pt}
\caption{\textbf{Average results on the DyNeRF dataset~\cite{li2022neural}.}}
\label{tab:dynerf_nvidia_merge}
\resizebox{1.0\linewidth}{!}{
\begin{tabular}{l|ccc}
\toprule
& \multicolumn{3}{c}{\textbf{DyNeRF}} 
\\
\cmidrule{2-4} 
\textbf{Methods} & \textbf{PSNR$\uparrow$} & \textbf{SSIM$\uparrow$} & \textbf{LPIPS$\downarrow$} 
\\
\midrule
HexPlane~\cite{cao2023hexplane}  & 15.33 & 0.5593 & 0.4514 \\ 
SC-GS~\cite{huang2024sc}     & 18.77 & 0.7359 & 0.2310  \\ 
D-GS~\cite{yang2024deformable}      & \trd 19.55 & \trd 0.7446 & \trd 0.2171 \\ 
MoDGS~\cite{qingming2025modgs}      & \snd22.64 & \snd0.8042 & \snd 0.1545 \\ 
\textbf{\prjName (ours)} & \fst 26.64 & \fst 0.8806 & \fst 0.1110 \\
\bottomrule
\end{tabular}}
\vspace{-1em}
\end{table}

\textbf{Evaluation Metrics.}
We evaluate rendering quality using three standard metrics for NVS: PSNR for pixel-wise fidelity, SSIM for structural similarity, and LPIPS for perceptual distance in deep feature space. All metrics are computed on cameras 5 and 6 to evaluate generalization to entirely unseen viewpoints.

\textbf{Pose Alignment.}
The depth maps used to generate Gaussian splats are produced in metric scale by off-the-shelf monocular depth estimation networks. However, the DyNeRF camera poses are in a scale set by COLMAP. Therefore, we perform a global scale-and-shift alignment between the depth maps and the COLMAP reconstruction. This brings both into the same coordinate system, ensuring geometric consistency and ensuring that comparisons reflect reconstruction quality rather than pose mismatches.
It is worth highlighting that this procedure is meant for evaluation purposes only, and it is not required for in-the-wild NVS.

\begin{figure}[t]
    \centering
    \includegraphics[width=0.95\linewidth]{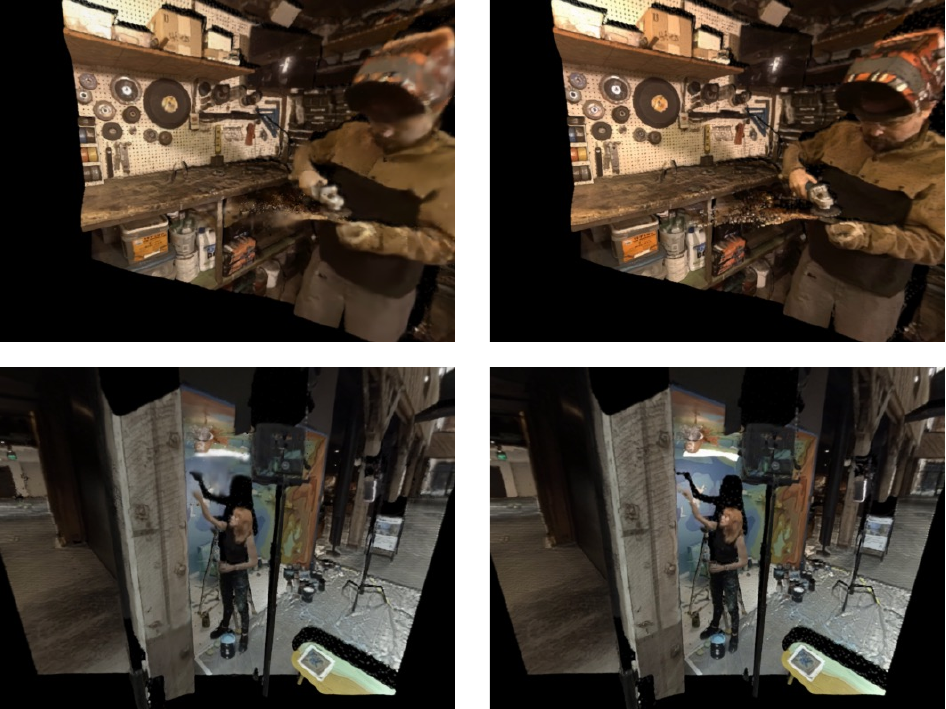}
    
    \caption{\textbf{Qualitative results on sequences from the Light Field Video dataset \cite{broxton2020immersive}.} 
Each pair compares a frame rendered by \prjName (left) with the corresponding ground truth image from one of the dataset's camera poses (right).}

    \label{fig:lfv_results}
\end{figure}

\subsection{Comparison with State-of-the-Art}

First, we compare \prjName with state-of-the-art approaches targeting the same scenario as ours.
Table \ref{tab:DyNeRF_perscene} presents the results on each individual scene in the DyNeRF dataset. We can observe how \prjName consistently outperforms all prior work on every scene and according to all evaluation metrics. 

Focusing on the PSNR metric, \prjName often achieves remarkable improvements over the second-best method, MoDGS, which frequently surpass the 2dB barrier -- up to more than 6dB on \texttt{sear\_steak}. 
Results averaged over the entire DyNeRF dataset are reported in Table \ref{tab:dynerf_nvidia_merge}, confirming a mean improvement of 4dB over MoDGS. 

Fig.~\ref{fig:lfv_results} reports a qualitative comparison on the Light Field Video dataset, where rendered frames are compared with the corresponding ground-truth images from one of the dataset cameras. 
These results indicate that \prjName can produce plausible renderings even when applied to different kind of video data.

\begin{table}[t]
\centering
\caption{\textbf{Ablation Studies.} Impact of the number of proxy images (A-D) and the depth refinement stage (E-F).}
\label{tab:ablation}
\renewcommand{\tabcolsep}{12pt}
\resizebox{1.0\linewidth}{!}{
\begin{tabular}{cl|ccc}
\toprule
& \multicolumn{3}{c}{\textbf{DyNeRF}} \\
\cmidrule{2-4} 
& \textbf{Methods} & \textbf{PSNR$\uparrow$} & \textbf{SSIM$\uparrow$} & \textbf{LPIPS$\downarrow$} \\
\midrule
(A) & 2 Synthetic Cameras & 23.77 & 0.8280 & 0.1921 \\ 
(B) & 4 Synthetic Cameras & \trd 26.08 & \trd 0.8780 & \trd 0.1166 \\ 
(C) & 8 Synthetic Cameras & \snd 26.15 & \snd 0.8790 & \snd 0.1141 \\ 
(D) & 18 Synthetic Cameras & \fst 26.64 & \fst 0.8806 & \fst 0.1110 \\

\midrule
(E) & w/o Refinement & \snd 26.58 & \snd 0.8763 & \snd 0.1165  \\
(F) & w/ Refinement & \fst 26.64 & \fst 0.8806 & \fst 0.1110 \\
\bottomrule
\end{tabular}}
\vspace{-1em}
\end{table}

\subsection{Ablation studies}

We conclude by assessing the impact of the different design choices behind \prjName. The outcome of this study is reported in Table \ref{tab:ablation}.

\textbf{Number of Proxy Images.} Rows (A) to (D) report the results achieved by increasing the number of proxy images generated during the initial stage. We can observe how two images already allow for obtaining high-quality renderings; however, a noticeable improvement can be achieved by increasing the number of proxy views to four. Adding further proxy images still contributes to better results, although with minor impact.

\begin{figure}[t]
    \centering
    \includegraphics[width=0.48\linewidth]{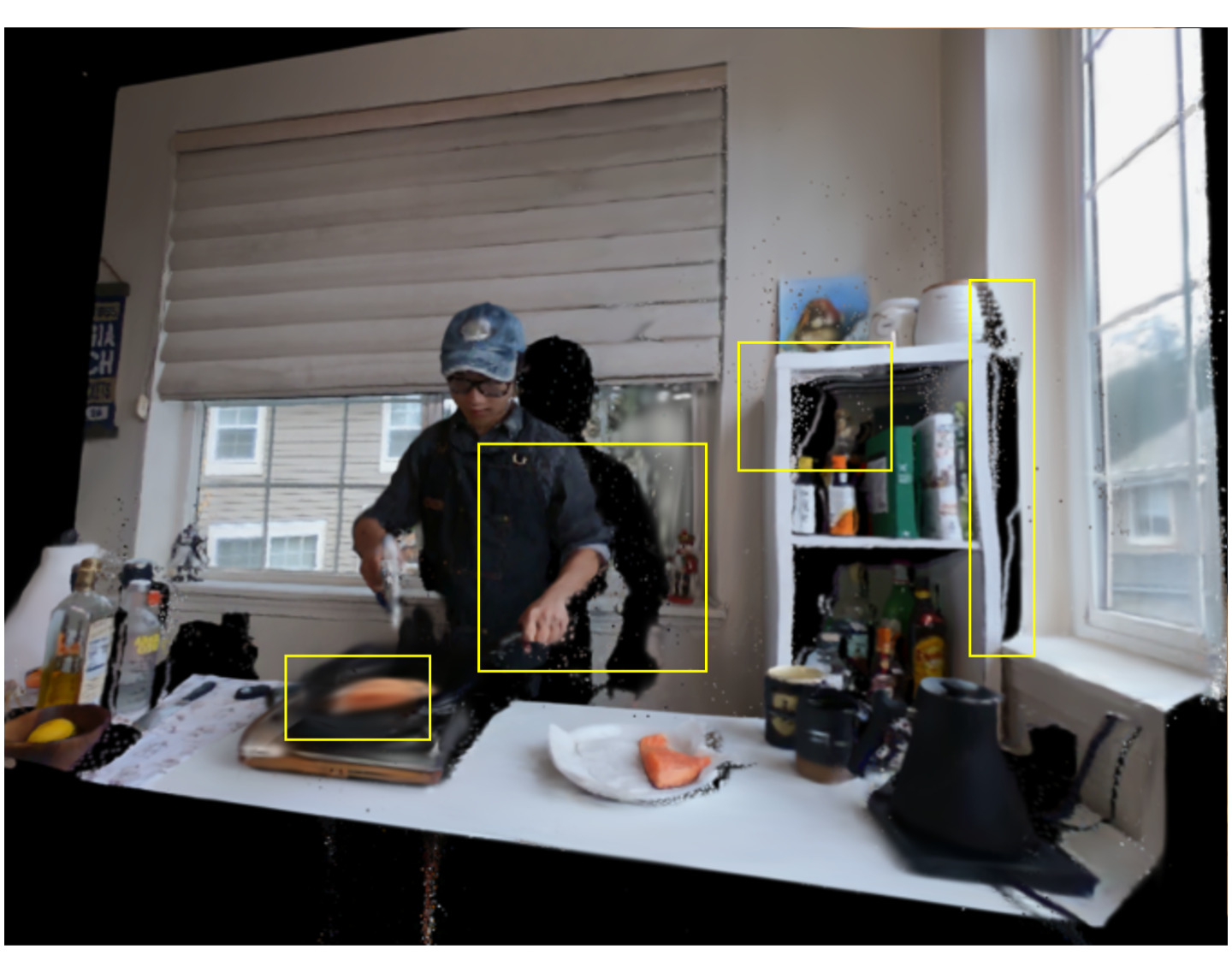}
    \includegraphics[width=0.48\linewidth]{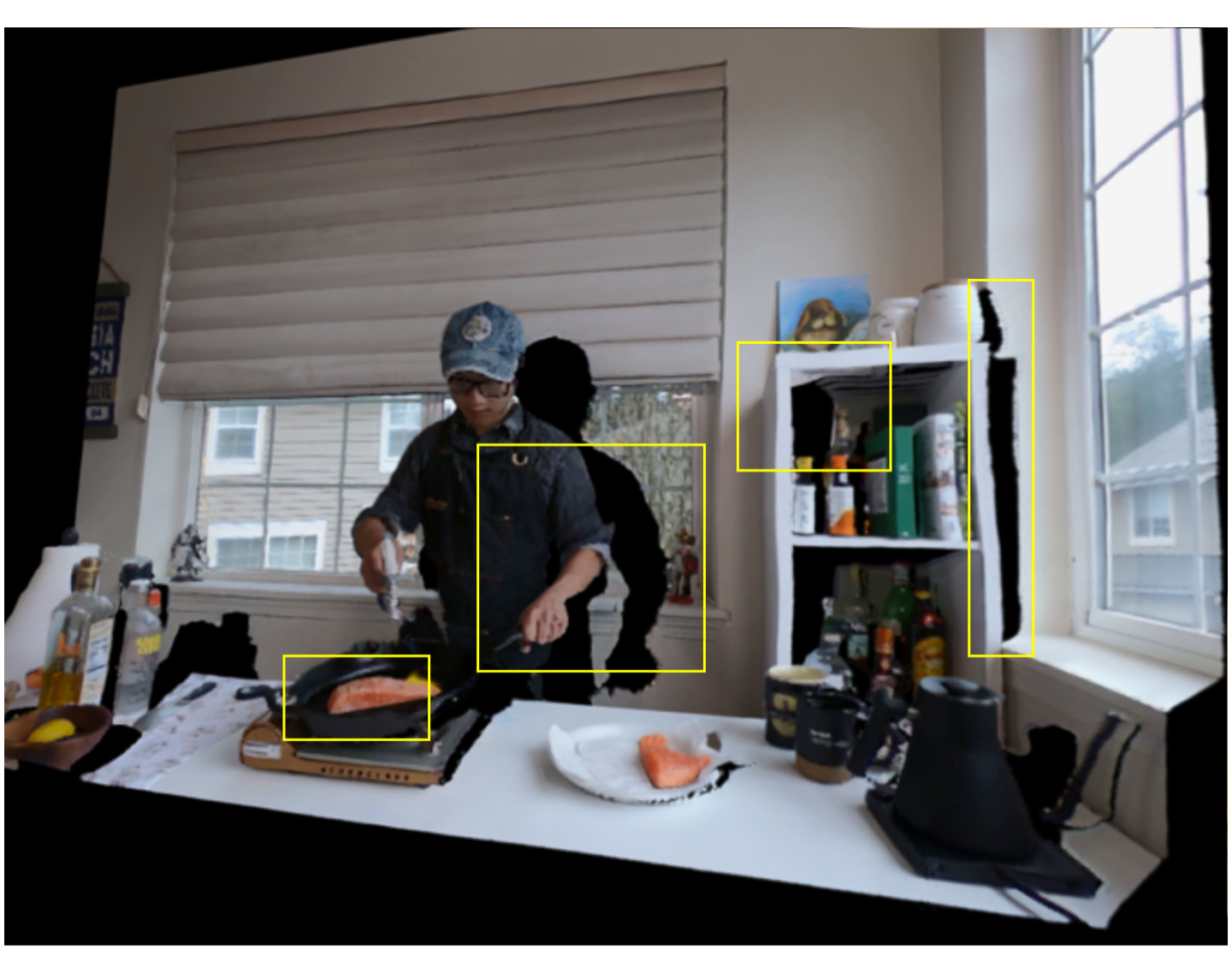}
   
    \caption{\textbf{Qualitative comparison on \textit{flame\_salmon\_1} sequence -- w/o Refinement (top) vs w/ Refinement (bottom).} Depth refinement allows for cleaner discontinuities and higher-frequency details (see the yellow boxes).}
    \label{fig:refinement}

\end{figure}

\textbf{Depth Refinement.} Rows (E) and (F) show the impact of depth refinement on the quality of the rendered images. Although not as impactful as the number of proxy images, we can appreciate how the refinement stage enables an overall improvement in the quality of the rendered images. This occurs near depth discontinuities in particular, as the refinement network excels at predicting sharp and accurate edges. Fig. \ref{fig:refinement} shows this comparison qualitatively, confirming how the refinement stage allows for rendering finer details and cleaner discontinuities. 

\section{Conclusion}

We presented \prjName, a novel dynamic 3D Gaussian Splatting framework tailored for monocular videos captured by static cameras. By leveraging depth-guided proxy image synthesis to generate synthetic multi-view supervision, our method overcomes the lack of geometric constraints inherent to static-camera settings. Extensive experiments on the DyNeRF benchmark demonstrate that \prjName surpasses existing approaches while requiring minimal external supervision, highlighting its potential for practical dynamic view synthesis in constrained capture scenarios.

\bibliographystyle{IEEEtran}
\bibliography{main}

\end{document}